\documentclass[letterpaper, 10 pt, conference]{ieeeconf}  

\IEEEoverridecommandlockouts                              

\usepackage{graphics} 
\usepackage{epsfig} 
\usepackage{amsmath} 
\usepackage{subfigure}
\usepackage{color}
\usepackage{hhline}
\usepackage{tabularx}
\usepackage[pagebackref=true,breaklinks=true,letterpaper=true,colorlinks,bookmarks=false]{hyperref}

\title{\LARGE \bf
Real-Time RGB-D based Template Matching Pedestrian Detection
}

\author{Omid Hosseini jafari and Michael Ying Yang}

\begin{document}

\maketitle
\thispagestyle{empty}
\pagestyle{empty}

\begin{abstract}
Pedestrian detection is one of the most popular topics in computer vision and robotics.
Considering challenging issues in multiple pedestrian detection, we present a real-time 
depth-based template matching people detector. In this paper, we propose different 
approaches for training the depth-based template. We train multiple templates for 
handling issues due to various upper-body orientations of the pedestrians and different 
levels of detail in depth-map of the pedestrians with various distances from the camera.
And, we take into account the degree of reliability for different regions of sliding 
window by proposing the weighted template approach.  Furthermore, we combine the
depth-detector with an appearance based detector as a verifier to take advantage of 
the appearance cues for dealing with the limitations of depth data. We evaluate our method 
on the challenging ETH dataset sequence. We show that our method outperforms the 
state-of-the-art approaches.
\end{abstract}

\section{INTRODUCTION}
\label{sec:intro}
Pedestrian detection is one of the most popular topics in computer vision. It has plenty of interesting applications in robotics, automotive safety, surveillance and autonomous vehicles, which makes it an attractive research topic \cite{Dalal05CVPR, DollarBMVC09ChnFtrs, DollarBMVC10FPDW, DollarPAMI14pyramids, Benenson2012Cvpr, Luo_2014_CVPR, paisitkriangkrai:strengthening, Ouyang13ICCV, Tian_2015_CVPR}. 
The performance of most tracking-by-detection approaches \cite{Andriluka08CVPR}, \cite{Park10ECCV}, \cite{Andriluka10CVPR}, \cite{HosseiniICRA14} is dependent on the pedestrian detector. In this paper, we address the problem of multiple people detection in the scenarios with a moving stereo/Kinect-based camera. The camera can be installed on a robot with 3-degrees of freedom (i.e., translation along $x$ and $z$ axes, and rotation around $y$ axis) or mounted on a helmet with 6-degrees of freedom (i.e., translation along and rotation around all axes).

One of the considerable challenges in pedestrian detection is the computational time complexity. Particularly, the detectors which are using appearance based features \cite{Tian_2015_CVPR, Dalal05CVPR, Ouyang13ICCV} have high computational cost. Computation of appearance based features can be accelerated by using GPU programming \cite{Benenson2012Cvpr}, \cite{Prisacariu09TR}. However, it is not preferred in robotics due to high energy consumption of GPU.

Another major challenge is occlusion handling. There are many cases which the pedestrians are very close to the camera and most of the lower parts of their body are outside the camera image plane. These cases are handled by depth based upper-body detectors \cite{Mitzel12BMVC}, \cite{HosseiniICRA14}. Although, the depth based people detectors perform well in challenging scenarios, they have some issues due to limitations of depth maps. Normally, depth maps for close objects are reliable but for far away objects the depth maps are noisy and this noise affects directly the shape of the objects. The other limitation of using only depth maps as input is the similarity of depth maps between some objects and the upper-body of human which leads to false positives.

Most of the state-of-the-art pedestrian detectors apply template matching approach \cite{Dalal05CVPR}, \cite{HosseiniICRA14}, \cite{Choi13PAMI}, \cite{Ouyang13ICCV}, \cite{Tian_2015_CVPR}. Handling various body orientations is one of the challenges when employing template matching detectors. Another concern is handling different levels of detail for pedestrians with different distances from the camera.

Our motivation in this paper is to improve the performance of the depth based template matching detectors considering aforementioned challenges. For this end, we have the following main contributions during this work. We propose different training approaches. We consider the reliable and unreliable regions inside the sliding windows, by training a weighted template. On the other hand, we take care of  different body orientations by introducing multiple orientation based templates. Furthermore, we deal with different levels of details in depth maps by training multiple templates for various distance ranges from the camera. Additionally, for dealing with depth map limitations such as noise and similar objects we propose to verify the resulting detections of depth based detector with an appearance based detector. Meanwhile, we aimed to keep whole the pipeline running real-time on a single CPU core. Finally, we evaluate our work with other state-of-the-art detectors on a challenging dataset ETH Zurich Sunnyday \cite{Ess09ICRA} and it outperforms the state-of-the-art approaches.

\subsection{Related Work}
Some of the pedestrian detectors use appearance based features \cite{Dalal05CVPR, DollarBMVC09ChnFtrs, DollarPAMI14pyramids, Benenson2012Cvpr}, which are generated from RGB channels of a monocular input image. These methods are applied on full-body of pedestrians and yields remarkable detection performance. Run-time complexity of these approaches is an issue for using them in robotics applications. The most complex part of these approaches is the extraction of appearance based features which are used to find the boundary and shape of the objects. Some of these approaches \cite{Benenson2012Cvpr} deal with this issue by implementing their works on GPU. 

With the aid of depth-maps which are obtainable from stereo images or Kinect-based cameras, extracting the boundary of objects is pretty faster and more accurate. Some approaches take advantage of this fact and propose the depth-based detectors \cite{Mitzel11BMVC}. Also, some other approaches combine the appearance based features and depth-maps and propose RGB-D based detectors \cite{HosseiniICRA14}, \cite{Munaro12IROS}. Hosseini \textit{et al.} \cite{HosseiniICRA14} combine a upper-body depth detector, for detecting the close range pedestrians, and a HOG based detector, for detecting the farther pedestrians. We used a different strategy for combining depth-based and RGB based detectors. We use an appearance based detector as a verifier to improve the performance of the depth-based detector. In this case, the appearance based features are computed only for the contents inside the detections which are verified. Therefore, our approach can be run on CPU in real-time.

\section{Overview}
\begin{figure}[t]
	\centering
	\includegraphics[width = \columnwidth]{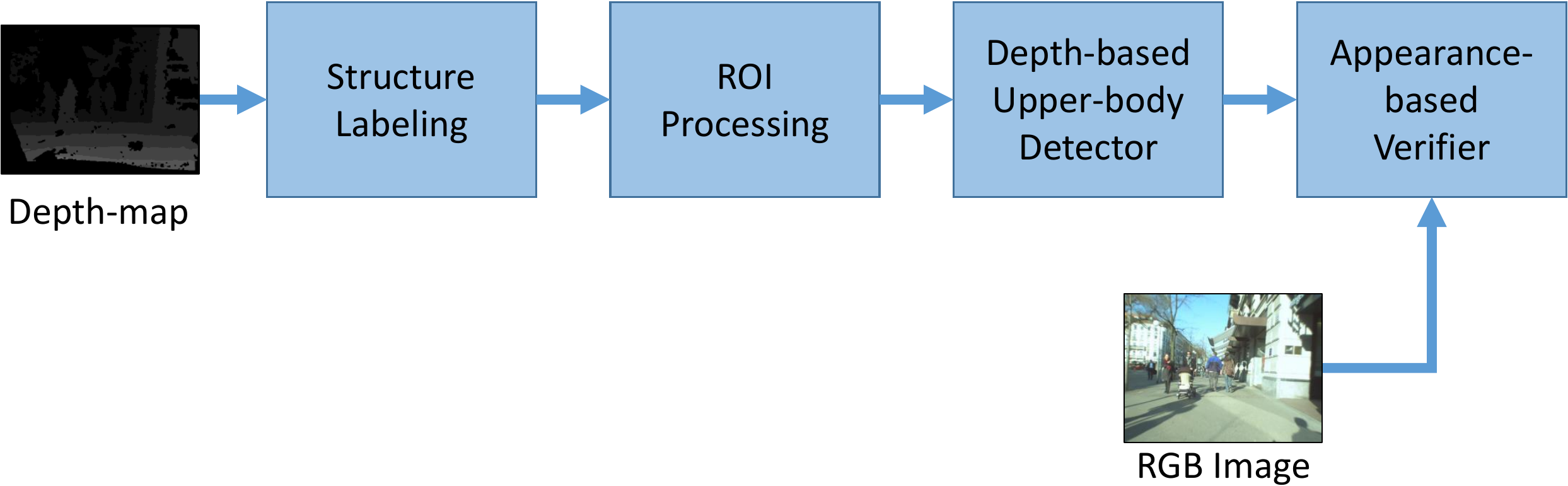}
	\caption{Pipeline of our RGB-D Pedestrian detector.}
	\label{fig:overview}
\end{figure}

The pipeline of our multiple pedestrian detector is shown in Fig. \ref{fig:overview}. The first three components (structure labeling, ROI processing and depth-based upper-body detector) are dependent only on the depth-map, while the verifier component uses the RGB image for refining the detections. 

\subsection{Structure Labeling}
First, the input depth-map is passed to the structure labeling component. The goal of this part is to remove the elevated structures (e.g. building and trees), estimate the ground plane and find the related 3D points to the objects. Estimation of the ground plane is only necessary when the camera is installed on a helmet, otherwise for the cases which camera is installed on a robot the ground plane is fixed. For estimating the ground plane, the 3D points, which are generated using depth-map, are projected to a rough ground plane considering the initial distance of the camera from the ground. Then, an occupancy map \cite{Ess09ICRA}, \cite{Badino07ICCV} is computed and the points in the bins with low densities are selected for estimation. The ground plane is estimated by applying RNASAC \cite{Fischler81ACM} plane fitting on the selected 3D points. After estimating the ground plane, the 3D points are projected to the estimated ground plane. Then, the ground plane is subdivided into a 2D grid. For each cell of the grid, a hight histogram is computed based on the hight of each 3D point inside the cell. The hight histogram contains 4 bins: \textit{ground plane}, \textit{object}, \textit{free space} and \textit{elevated structures}. In these scenarios, we assume a free space on top of the objects in the scene to avoid to label pedestrians, which are walking below hanging objects, as \textit{elevated structures}.

\subsection{Region of Interest Processing}
Afterwards, the 3D points labeled as object and the estimated ground plane are passed to the ROI processing component. The goal of this part is to segment the 3D points into hypothesis objects and for each object find the corresponding bounding-box on the image plane. For this end following \cite{Bajracharya09IJRS},\cite{Bansal10ICRA},\cite{HosseiniICRA14}, the input 3D points are projected to the ground plane and a 2D histogram is computed. Then, we find the connected components inside the resulting 2D histogram. Each connected component corresponds to a ROI. Finally, the ROI bounding-boxes on the image plane is computed by back projecting the 3D points inside each component.

\subsection{Depth-based Upper-body Detector}
\label{subsec:intro:detection}
The overall pipeline of the depth-based upper-body detector is based on \cite{HosseiniICRA14}. We significantly improve this approach by proposing three different training phases to handle the mentioned issues in Sec.\ref{sec:intro}. In the following we briefly overview this approach.

The depth-based upper-body detector is a depth-based template matching detector. It has two main phases: \textit{training} and \textit{detection}. In the training phase, given a training set $X$ of $N$ normalized upper-body annotations $\mathbf{x}$, the depth-based template is computed as follows:
\begin{equation}
\mathbf{t}=\frac{1}{N}\sum\limits_{i=1}^{N}\mathbf{x}_i=\bar{\mathbf{x}}.
\end{equation}
The annotations are cropped from the depth-maps of the training frames and resized into a fixed size ($150\times 150$). The trained template is used during the detection phase.

For each ROI bounding-box inside a frame, we cropped the depth map from the input frame and rescale it based on the ratio of hight of the bounding-box to hight of the template. The ROI is normalized using the median of its centering points. Afterwards, the contour of ROI is extracted by simply finding the maximum $y$ value in each $x$ position, and the local maxima of the resulting contour is extracted. Then, difference between the template and the ROI is computed only on local maxima positions. The Euclidean distance is used as the distance measurement.

\begin{figure}[t]
	\centering
	\includegraphics[width = 0.7\columnwidth]{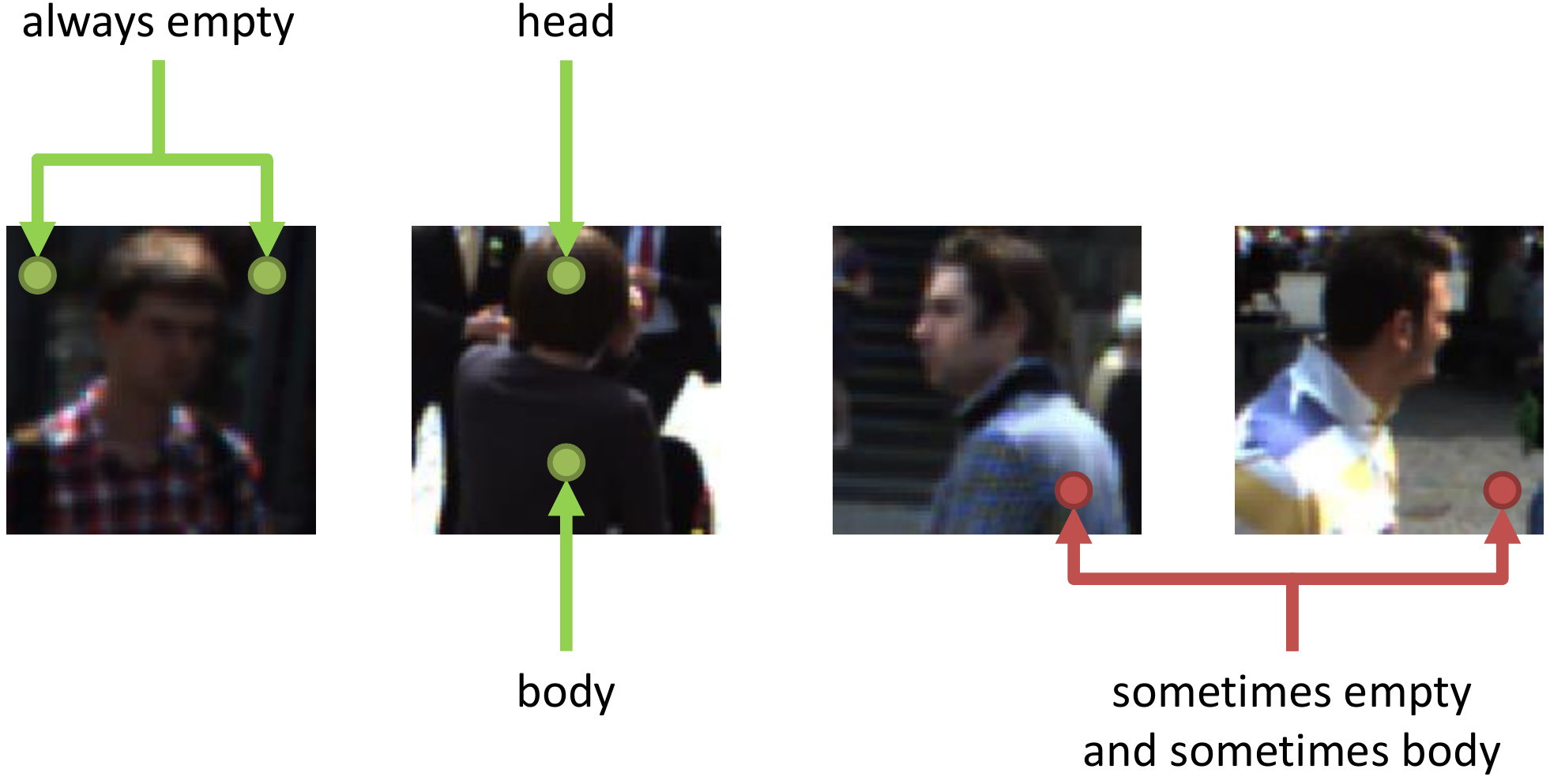}
	\caption{Each part of a bounding-box containing a human upper-body has different importance.}
	\label{figure:template_area_importance}
\end{figure}

\section{Weighted Template}
In depth-based template matching approaches \cite{Mitzel12BMVC,HosseiniICRA14}, a depth template is computed by averaging over upper-body annotations. This template is compared with the contents of a sliding window over depth-map of test image. These approaches assume different regions inside an annotated upper-body bounding-box have the same importance, however in real data different regions have different importance. As it is shown in Fig. \ref{figure:template_area_importance}, the middle-top and middle-bottom regions of the bounding-box always contain the body parts and the right-top and left-top corners are always empty. Hence, during the detection phase these regions are expected to have the consistent state inside the test sliding window. In contrast, the right-bottom and left-bottom corners do not have the same states all the time. These regions have less importance and they are less reliable during the detection phase. Therefore, high difference between template and the sliding window inside unreliable regions can cause false negative detections. On the other hand, low difference inside important regions can lead to false positive detections. We propose to weigh the different regions of the distance matrix (between the template and the sliding window) based on its importance. For this end, we train our template jointly with a weight matrix and change the distance measurement to consider the trained weight during detection phase.

\begin{figure}[t]
	\centering
	\includegraphics[width = 0.7\columnwidth]{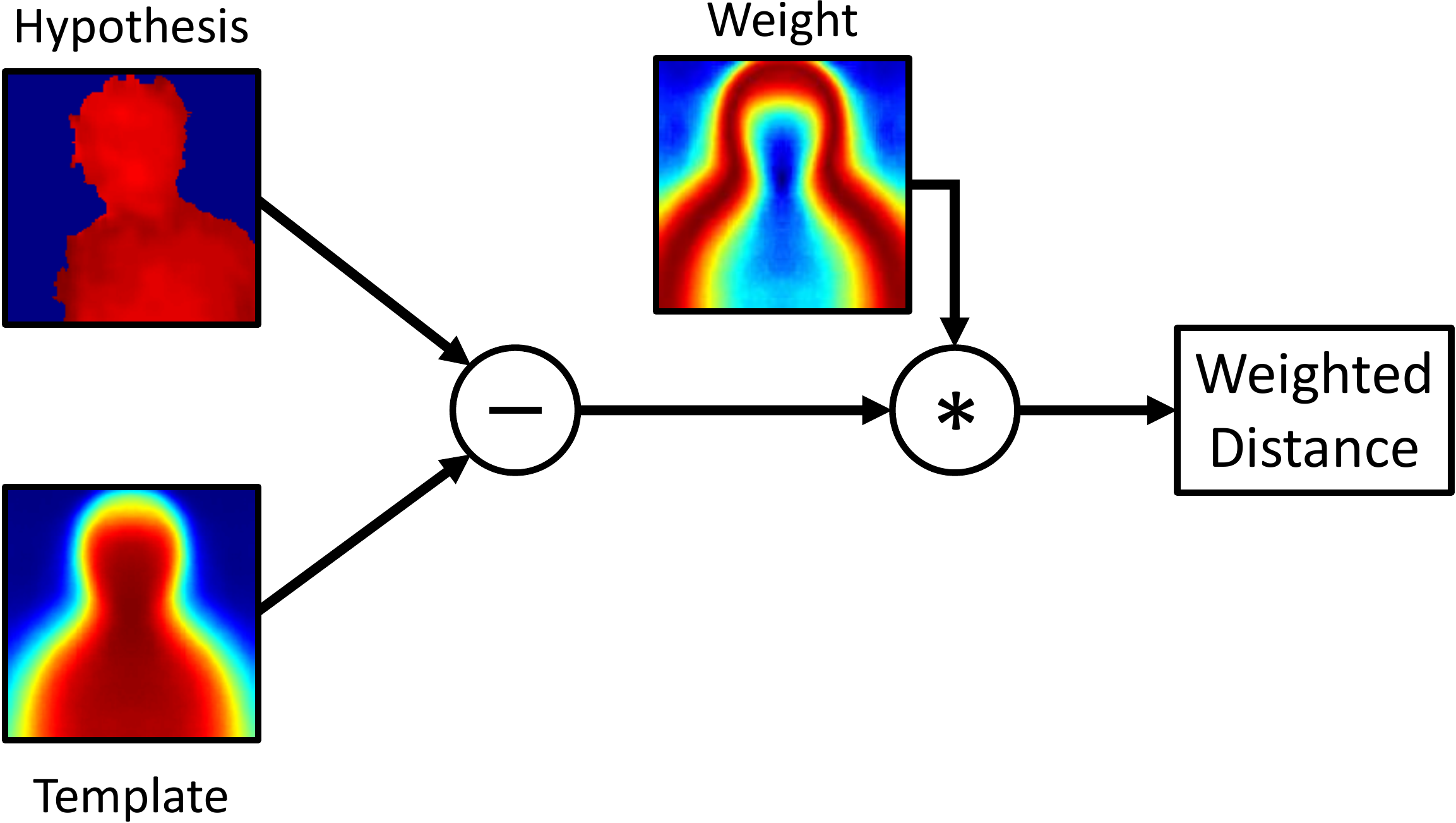}
	\caption{Computing the weighted distance. The distance between template and the sliding window is computed and multiplied by the weight matrix.}
	\label{fig:weight_detection}
\end{figure}

\subsection{Training of Weighted Template}
First, we define the weighted distance measurement $d$ as follows:
\begin{equation}
\label{eq:weighted-distance}
d=(\mathbf{t}-\mathbf{x})^2\mathbf{w},
\end{equation}
where $\mathbf{t}$ is the template, $\mathbf{x}$ is the content of the sliding window and $\mathbf{w}$ is the weight matrix with the same size as template. Then, we define an energy function $E$ in order to train the template and the weight matrix by minimizing the energy as follows:
\begin{eqnarray}
E=\sum\limits_{i=1}^{N}\mathbf{w}(\mathbf{t}-\mathbf{x}_i)^2+\frac{1}{\Arrowvert \mathbf{w}\Arrowvert}
\end{eqnarray}
\[\frac{\partial E}{\partial{\mathbf{t}}}=\sum\limits_{i=1}^{N}2\mathbf{w}(\mathbf{t}-\mathbf{x}_i)=0 \Longrightarrow \mathbf{t}=\frac{1}{N}\sum\limits_{i=1}^{N}\mathbf{x}_i=\bar{\mathbf{x}} \]
\[\frac{\partial E}{\partial \mathbf{w}}=\sum\limits_{i=1}^{N}(\mathbf{t}-\mathbf{x}_i)^2-\frac{1}{\mathbf{w}^2}=0 \Longrightarrow \mathbf{w}=\sqrt{\frac{N}{\sum\limits_{i=1}^{N}(\bar{\mathbf{x}}-\mathbf{x}_i)^2}}=\frac{1}{\mathbf{\sigma}},\]
where we add $\frac{1}{\Arrowvert w\Arrowvert}$ to the energy as a penalty for the low weights to avoid the zero weights. As it is shown, the energy minimization has a closed-form solution. Thus, the template is computed as in \cite{HosseiniICRA14} and the weight $\mathbf{w}$ is the reverse of the standard deviation of the training samples.

\subsection{Detection with Weighted Template}
During detection phase the wighted distance measurement (Fig. \ref{fig:weight_detection}) is used in the same approach as we explained in Sec.~\ref{subsec:intro:detection}.

\section{Multiple Templates}
Depth-based template matching upper-body detectors \cite{Mitzel12BMVC,HosseiniICRA14}, which use single-template for detecting the pedestrians, have following limitations:
\begin{enumerate}
	\item 
	A single-template mostly covers the front/rear side of the pedestrians upper-body since the pedestrians in the training-set are mostly walking towards or in the direction of the camera. Therefore, the pedestrians who appear with the left/right side in front of the camera will be penalized more with the single-template.
	
	Considering this problem, we can train more than one template based on the orientation (shape) of the training samples. For this end, we propose the \textit{orientation-based multiple templates} approach.
	\item The pedestrians with different distances from the camera have different levels of detail in their depth data. When only one template is generated from the training samples with the different distances from the camera, the resulting template will have an average detail. Therefore, this single-template penalizes the far away pedestrians due to their less details. Also, it penalizes the nearby pedestrians due to their more details in comparison to template.
	
	To solve this issue, we propose to learn \textit{distance-based multiple templates}.
\end{enumerate}
In the following, we will explain the training and the detection phases of orientation-based and distance-based multiple templates.

\begin{figure}[t]
	\centering
	\includegraphics[width = 0.7\columnwidth]{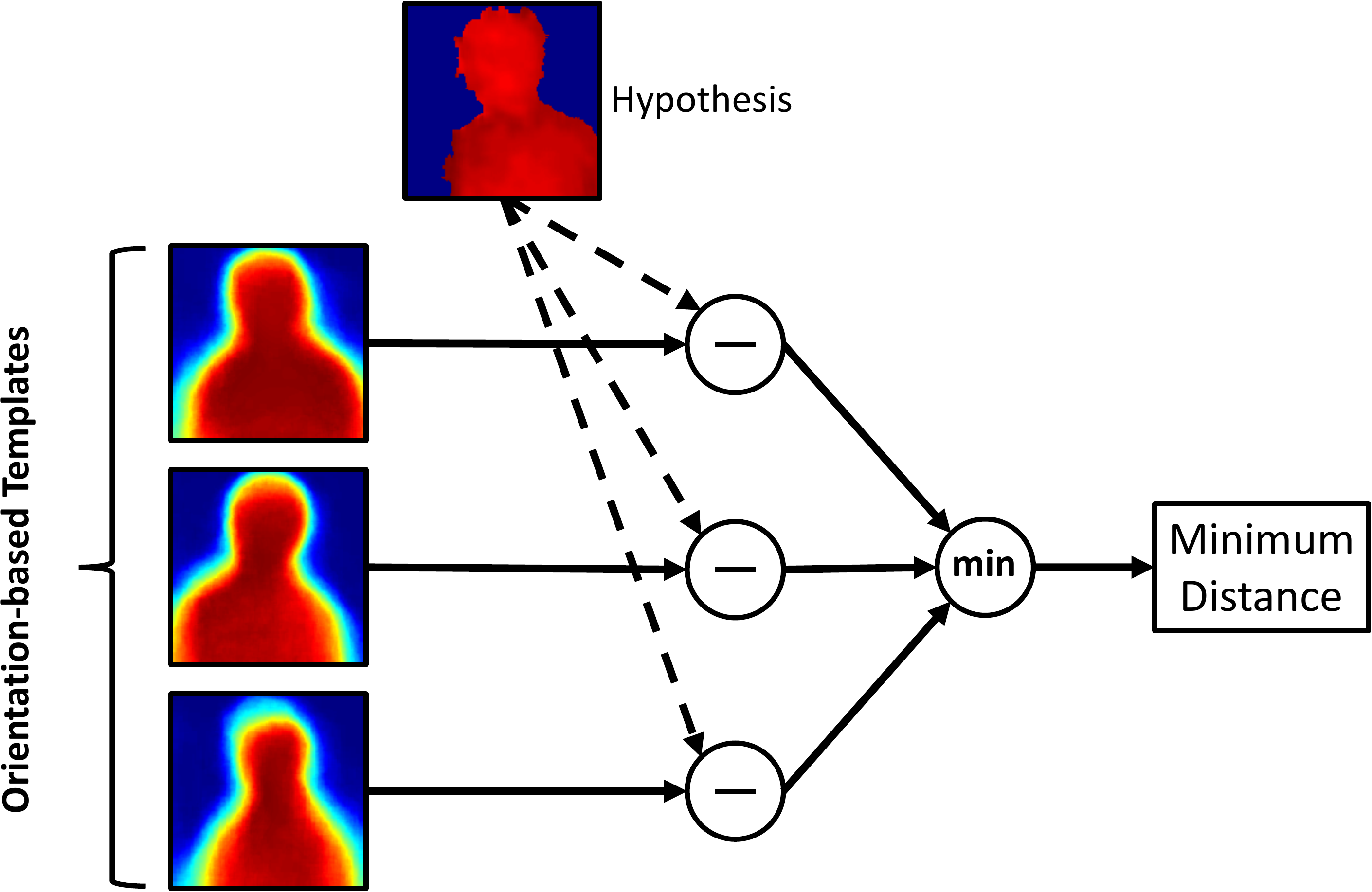}
	\caption{Computing the distance between the sliding window and multiple orientation-based templates.}
	\label{fig:multi-ori-detection}
\end{figure}

\subsection{Training of Multiple Templates}
Given a set of training samples $\mathbf{X}=\{\mathbf{x}_1,\mathbf{x}_2,...,\mathbf{x}_N\}$, our goal is to compute a set of templates $\mathbf{T}=\{\mathbf{t}_1,\mathbf{t}_2,...,\mathbf{t}_K\}$. The training phase of multiple templates has two steps:
\begin{enumerate}
	\item Clustering the training samples into $K$ subsets:
	\[\mathbf{X}_c=\{\mathbf{x}_{c1},\mathbf{x}_{c2},...,\mathbf{X}_{cN_c}\},\space N=\sum\limits_{c=1}^{K}N_c.\]
	\item Computing a template for each cluster by averaging over training samples inside their corresponding subset:
	\begin{equation}
	\mathbf{t}_c=\frac{1}{N}\sum\limits_{i=1}^{N_c}\mathbf{x}_c.
	\end{equation}
\end{enumerate}
The second step is the same for both orientation-based and distance-based approaches, but the clustering step is different for each approach.

\noindent
\textbf{Clustering for Orientation-based Templates:} We cluster the training data using k-means clustering method. One of the essential factors for orientation-based templates is the number of templates (clusters). For finding the best number of clusters, we compute \textit{silhouette scores} \cite{Rousseeuw1987SGA} of each clustering with different number of clusters. Silhouette score for each object, $i$, is computed as follows:
\begin{equation}
s(i)=\frac{b(i)-a(i)}{\max\lbrace a(i),b(i)\rbrace}
\label{eq_sil}
\end{equation}
where $a(i)=\frac{1}{N_{C_i}}\sum_{j\in C_i}dist(i,j)$ is the average of within-cluster distances of object $i$ and $b(i)=\min_{i\notin C_j}\lbrace dist(i,C_j)\rbrace$ is the average of between-cluster distances of object $i$. The scores are in the range $[-1,1]$ and objects with scores close to $1$ are the well-clustered objects. Finally, the clustering score is computed by averaging over the scores of the clustered objects. The quantitative evaluation results for different number of clusters is provided in Sec.\ref{sec:results}.

\noindent
\textbf{Clustering for Distance based Templates:} First, we define three distance ranges from the camera, $\{[0,4)m$, $[4,7)m$, $[7,+\infty)\}$. Then, we cluster training samples based on their distances from the camera. The median of depths inside a small bounding box in the middle of each annotation is used as the distance of the annotation from the camera.

\begin{figure}[t]
	\centering
	\includegraphics[width = 0.7\columnwidth]{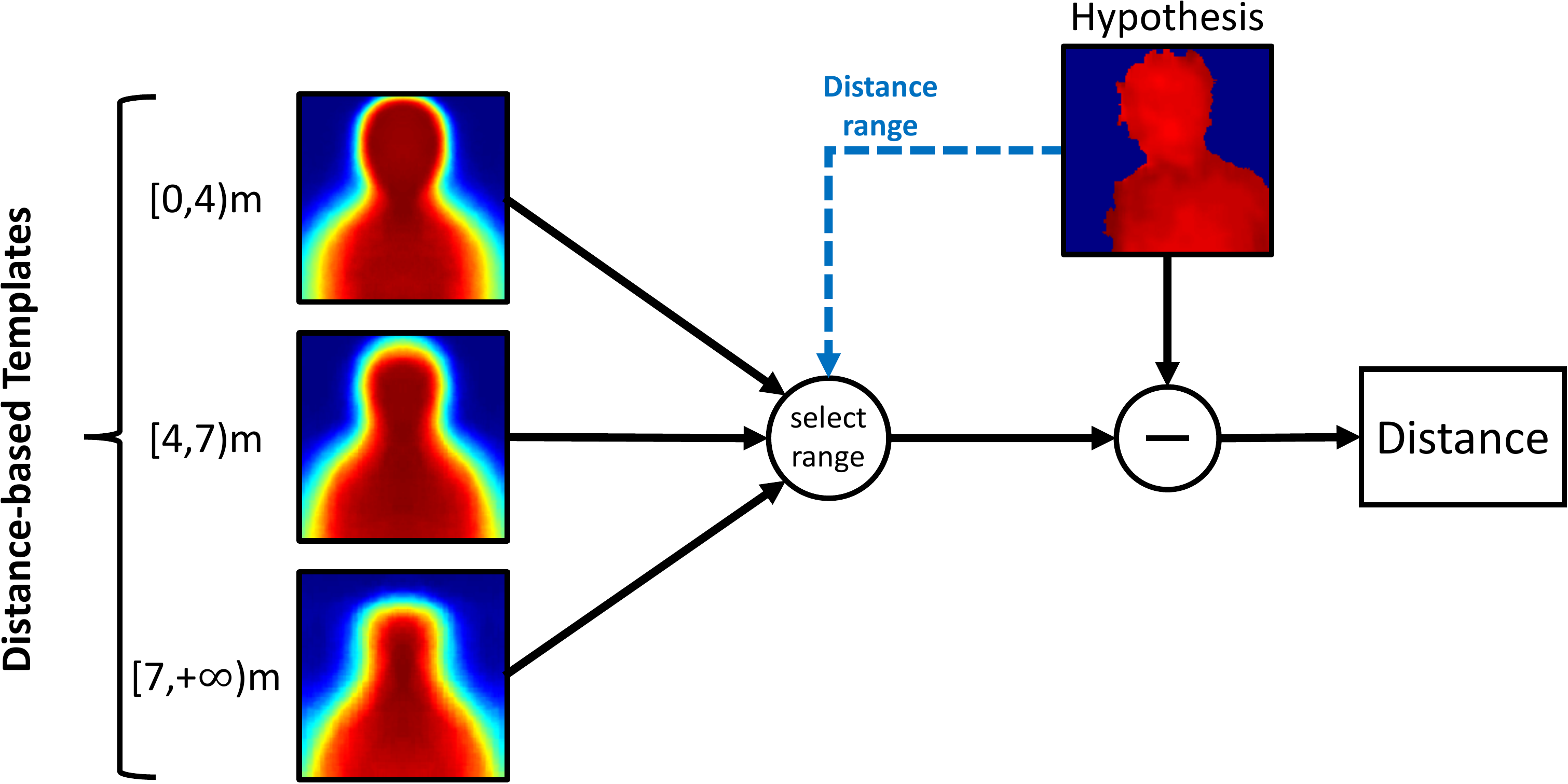}
	\caption{Computing the distance between the sliding window and its corresponding distance-based template.}
	\label{fig:multi-dist-detection}
\end{figure}

\subsection{Detection with Multiple Templates}
The detection pipeline is the same as we explained in Sec.~\ref{subsec:intro:detection} except the distance measurement part.

\noindent
\textbf{Orientation-based templates:} First, the distances between the sliding window and all the templates is computed. Then, the minimum distance is selected as the final distance for computing the score of the sliding window, as it is shown in Fig. \ref{fig:multi-ori-detection}.

\noindent
\textbf{Distance-based templates:} Based on the distance of the object inside the sliding window, the corresponding distance-based template is selected. Afterwards, the distance between the sliding window and the selected template is computed (Fig. \ref{fig:multi-dist-detection}).

\section{Appearance-based Verifier}
Depth based pedestrian detectors such as \cite{Mitzel12BMVC}\cite{Munaro12IROS}\cite{HosseiniICRA14} have good performance in real-time.  However, depth data has following limitations:
\begin{enumerate}
\item The depth map of the objects far away from the camera is mostly noisy and unreliable. These noisy depth maps can cause false negatives.
\item There are some objects with similar depth map as the human upper-body. Normally, these objects cause false positives.
\item Depth map is also unreliable when we are using a kinect-based camera and there is a strong light on the object. In such cases, mostly the head of the pedestrian will be removed and it does not detect (false negative).
\end{enumerate}
For solving these issues, we have to use appearance based features. The main problem of appearance based detectors is their high computational cost. The bottleneck of these approaches is the computation of the complex appearance based features for sliding windows in different positions with different scales.

In this paper we apply an appearance based pedestrian detector only as a verifier to verify the resulting detections, generated by the upper-body detector, which are not reliable.

First, we should find the unreliable detections. It is done by defining two thresholds as follows:
\begin{equation}
f(s)=\left\{
\begin{array}{llll}
\text{\textcolor{red}{Rejected}} &,&s&<th_{hard} \\[6pt]
\text{\textcolor{yellow}{Unreliable}} &,th_{hard}<&s&<th_{soft} \\[6pt]
\text{\textcolor{green}{Reliable}} &,th_{soft}<&s,
\end{array}
\right.
\label{eq:score_function}
\end{equation}
where $s$ is the score of the detection, $th_{hard}$ is the hard threshold and $th_{soft}$ is the soft threshold. The hard threshold is used to reject the bounding boxes with low scores. The detections with scores between hard and soft thresholds are labeled as unreliable and they should be verified. The main task of the verifier is to change the score of each unreliable detections by applying an appearance based detector on and around the corresponding bounding-box.

\begin{figure}
\centering
\includegraphics[width=\columnwidth]{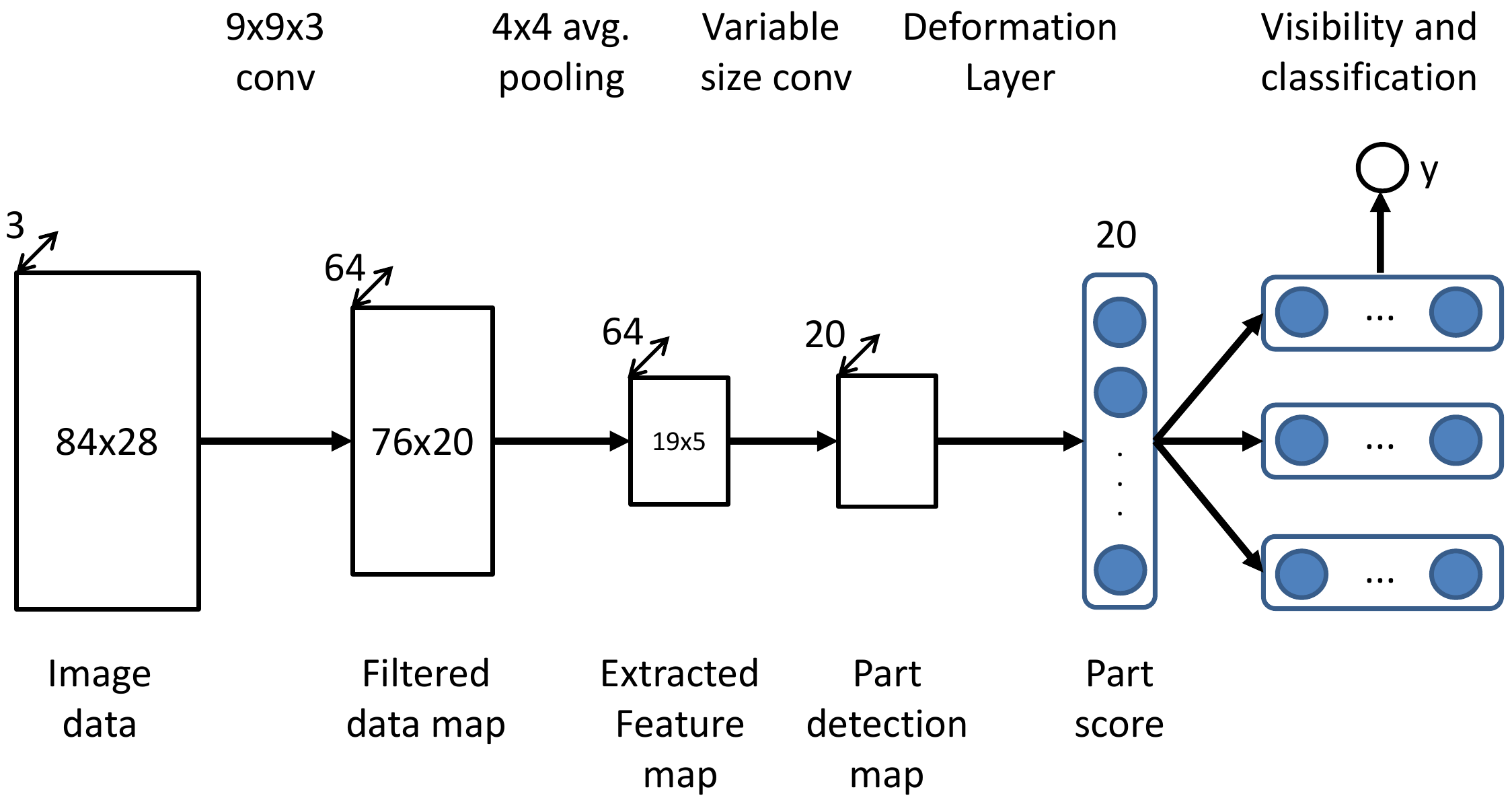}
\caption{Overview of the jointDeep \cite{Ouyang13ICCV} model.}
\label{fig:jointdeepmodel}
\end{figure}

In this paper, we use jointDeep pedestrian detector introduced by Ouyang \textit{et al.} \cite{Ouyang13ICCV} as the verifier. Fig.~\ref{fig:jointdeepmodel} shows the pipeline of the jointDeep model. In this model, feature extraction, a part deformation model, an occlusion model and classification are jointly learned on a deep model. This network contains two convolutional layers and one average pooling layer.

A detection window with the fixed size of $84\times 28$ is cropped from RGB image and converted to YUV color space. As an input, three $84\times 28$ channels will be generated. First channel is the Y-channel. Second channel is generated by concatenating three YUV channels which are resized to half. Third channel is generated by concatenating gradient magnitude of YUV channels and the maximum magnitudes of three of them.

Then, 64 filtered data maps are computed by convolving three input channels with $9\times 9\times 3$ filters. Afterwards, 64 feature maps are extracted by average pooling the filtered data maps using $4\times 4$ kernels and $4\times 4$ subsampling steps.

The 20 part detection maps have different sizes, since they are obtained by convolving the feature maps with part filters with variable sizes to model the occlusion. Next, part scores are computed for each part in the deformation layer. Finally the visibility reasoning of the obtained scores is used for computing the final score of sliding window, for more details see \cite{Ouyang13ICCV}.

As a verifier, we generate some bounding-boxes around the unreliable detections with the same ratio as $84\times 28$ (this is done since we might lose the head or shoulders of the upper-body in the obtained unreliable bounding-boxes due to unreliable depth maps as we discussed before). Afterwards, we extract these bounding-boxes from the RGB image, resize it to $84\times 28$ and pass it to the jointDeep detector. Finally, based on the obtained detection score from the verifier, we accept or reject the detection.

\begin{figure}
\centering
\includegraphics[width=0.9\columnwidth]{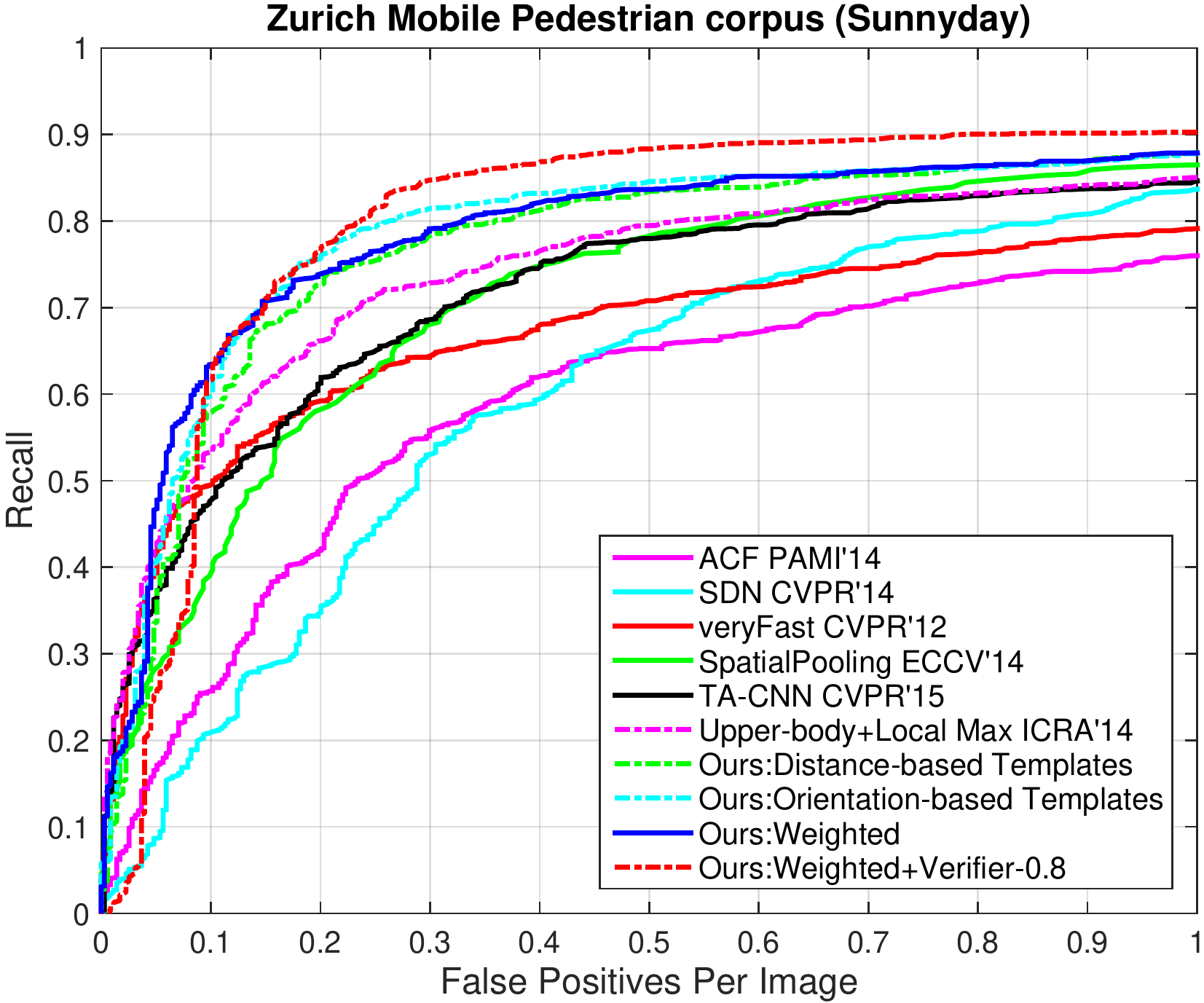}
\caption{Overal results on the ETH Zurich dataset \cite{Ess09ICRA}. We compare our proposed approaches with other state-of-the-art approaches. \textit{Our approaches}: depth-based upper-body detector using weighted template, orientation-based multiple templates and distance-based multiple templates without the appearance based verifier component, and depth-body detector using weighted template with jointDeep \cite{Ouyang13ICCV} as the appearance based verifier. \textit{State-of-the-art approaches}: ACF \cite{DollarPAMI14pyramids}, SDN \cite{Luo_2014_CVPR}, veryFast \cite{Benenson2012Cvpr}, SpatialPooling \cite{paisitkriangkrai:strengthening}, TA-CNN \cite{Tian_2015_CVPR} and upper-body+LocalMax \cite{HosseiniICRA14}. Our approaches outperform the state-of-the-art detectors.}
\label{fig:overal-results}
\end{figure}

\section{EXPERIMENTAL RESULTS}
\label{sec:results}
\subsection{Dataset}
For training the template(s) with the proposed approaches, the training dataset must fulfill the following requirements:
\begin{enumerate}
	\item Including well aligned annotations. This factor is really important specially for weighted template, since the scale and position of the pedestrians upper-bodies inside the annotation bounding-box must be the same for all the training samples to train a reliable importance region weight matrix.
	\item For training the distance-based multiple templates, sufficient number of annotations are needed in each distance ranges from the camera.
	\item In order to train orientation-based multiple templates, annotations with different orientations and in sufficient amount are needed.
\end{enumerate}
The templates are trained using 835 normalized $150\times 150$ depth annotations which are fulfilled the above requirements.

\begin{figure*}[t]
	\centering
	\subfigure[]{
		\includegraphics[width=0.9\columnwidth]{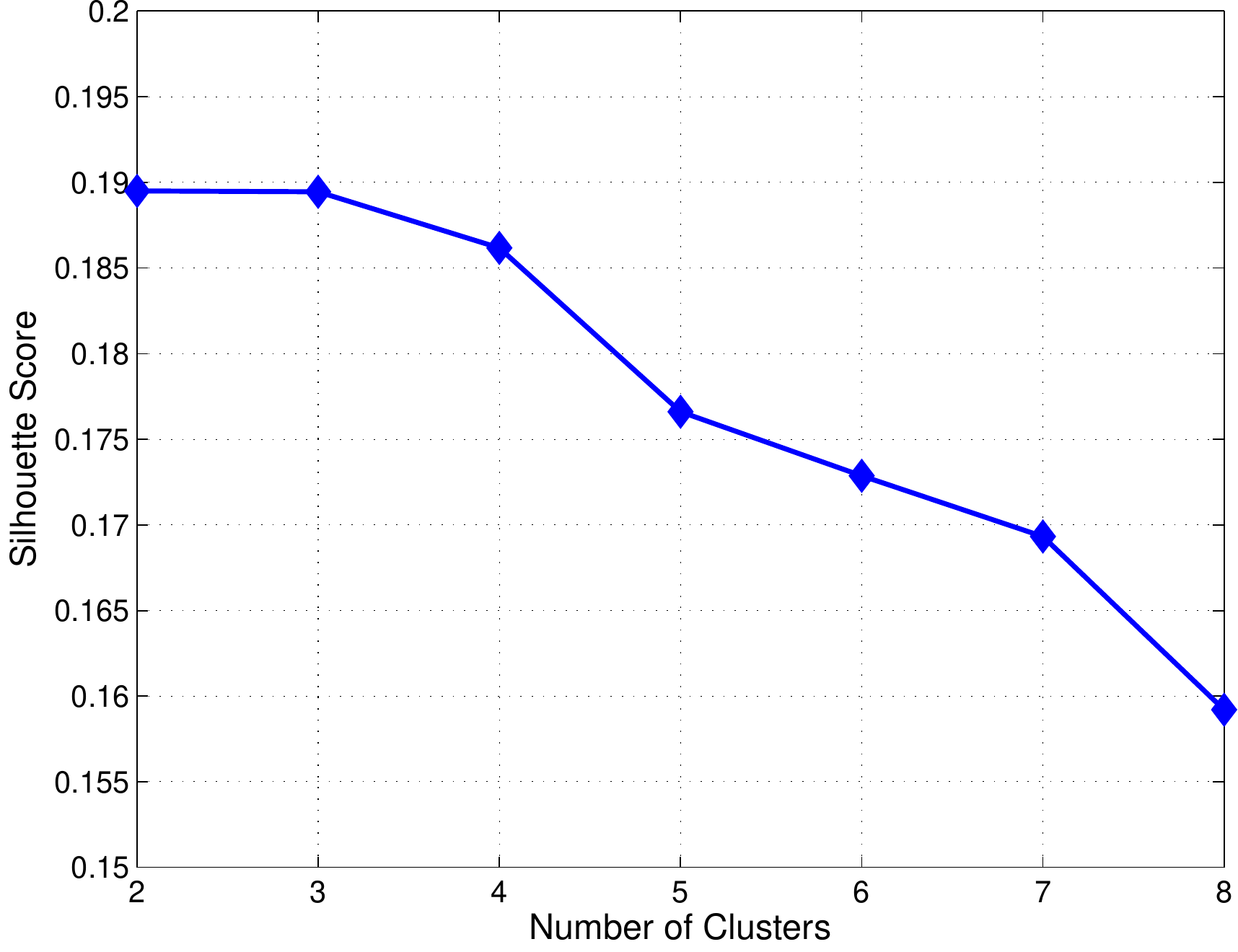}
	}
	\subfigure[]{
		\includegraphics[width=0.9\columnwidth]{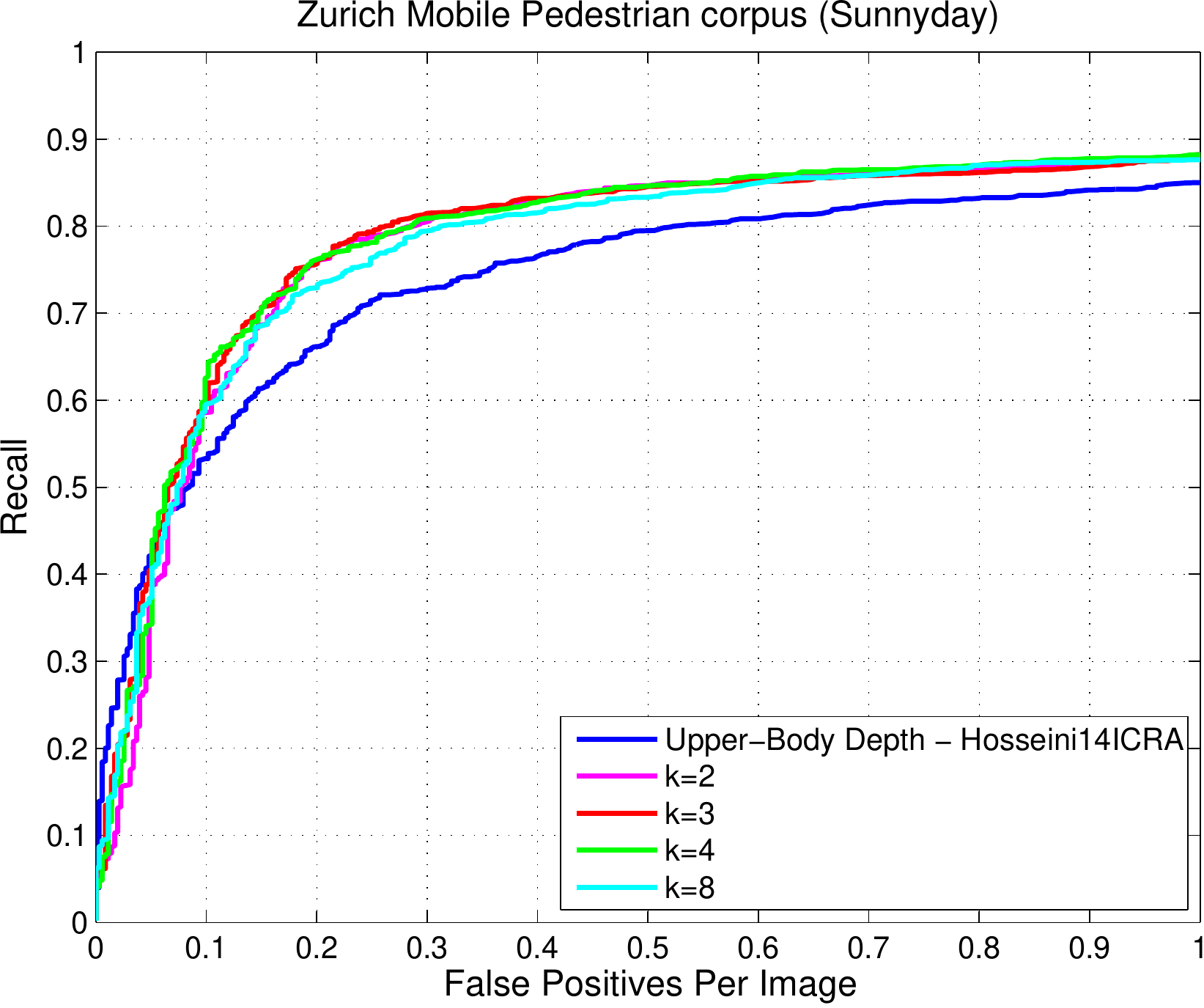}
	}
 \caption[]{(a) Illustration of Silhouette scores for clustering of training dataset with different number of clusters (higher scores are better). (b) Evaluation of orientation-based multiple templates with different number of templates on the ETH Zurich Sunnyday dataset \cite{Ess09ICRA}.}
  \label{fig:multi_orientation_clusters}
\end{figure*}

\subsection{Evaluation}
We evaluate our approaches on the ETH Zurich mobile pedestrian corpus Sunny Day dataset \cite{Ess09ICRA}. For obtaining the quantitative evaluations, the overlap of the detections and the annotations is measured. Then, the recall over false positives per image is plotted.
Fig.~\ref{fig:overal-results} shows the experimental results of our approaches and other state-of-the-art pedestrian detectors. We evaluate the performance of the depth-based upper-body detector with our proposed weighted template, orientation-based multiple templates and distance-based multiple templates with and without the appearance based verifier.

For training the distance-based multiple templates, we define three distance ranges, $\{[0,4),[4,7),[7,+\inf)\}$, and cluster the training data sets based on these distance ranges. As shown in Fig. \ref{fig:overal-results}, the performance of the upper-body detector is increased by $5\%$ when the distance-based templates are used instead of the single-template \cite{HosseiniICRA14}.


\paragraph*{Number of templates}
Fig. \ref{fig:multi_orientation_clusters}(a) shows the silhouette clustering scores when we cluster the training data with different number of clusters. As its is shown, clusterings with 2 and 3 clusters have the best score. We also generate multiple orientation-based templates with different number of templates and evaluate the performance of detector using them. Fig. \ref{fig:multi_orientation_clusters}(b) shows the quantitative evaluation of our proposed orientation-based multiple template approach with different number of templates and also compare them with the performance of \cite{HosseiniICRA14}. As it is shown, the orientation-based multiple template approach outperforms the single template approach. On the other hand, Fig. \ref{fig:multi_orientation_clusters}(b) shows that the orientation-based approach has its best performance with $K=3$ templates. For comparing the performance of the orientation-based multiple templates with other approaches, we use $K=3$ orientation-based templates. As shown in Fig. \ref{fig:overal-results}, it outperforms the single depth-based template approach \cite{HosseiniICRA14} and other state-of-the-art appearance based detectors.

\begin{figure}
\centering
\includegraphics[width=0.9\columnwidth]{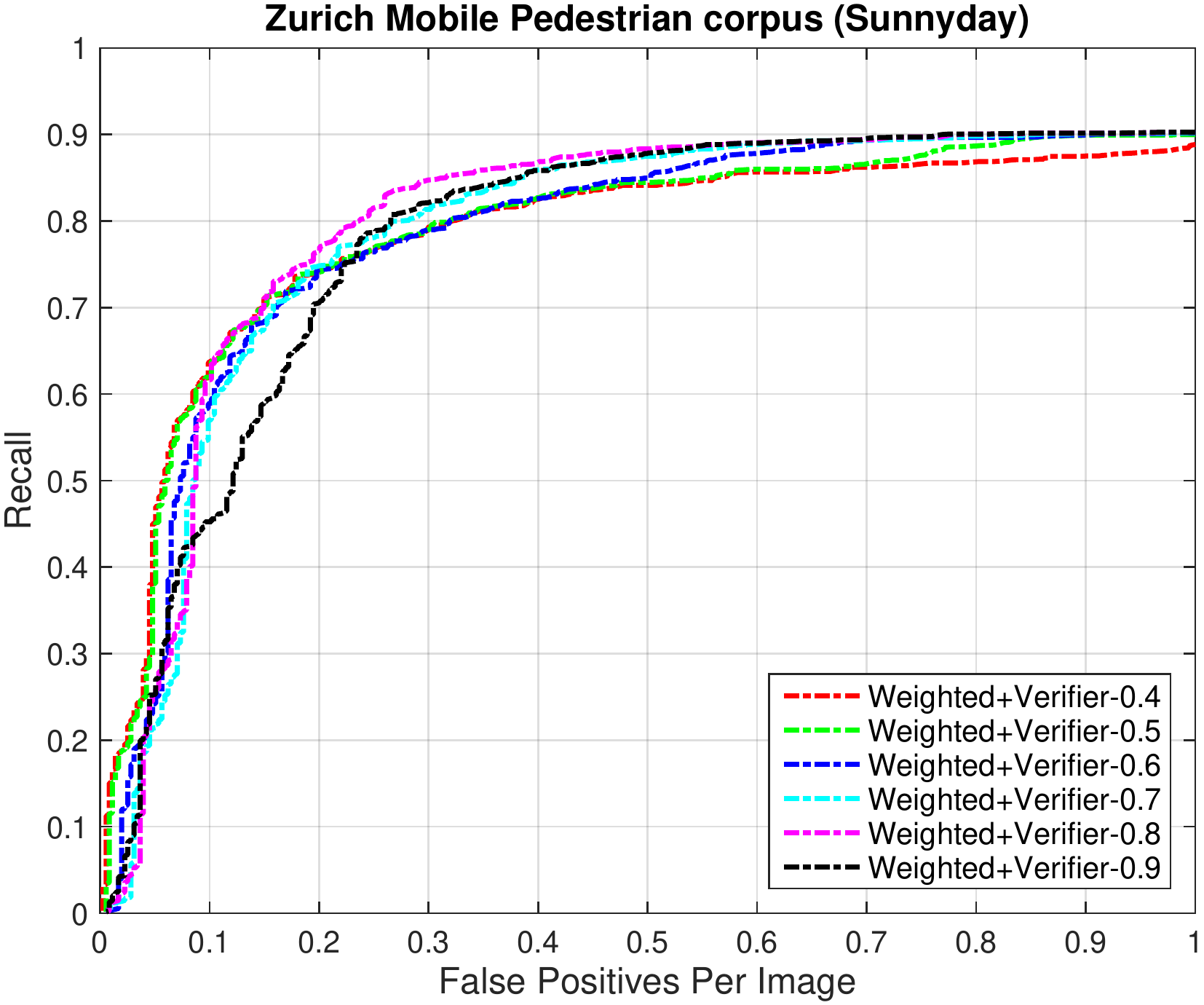}
\caption{Detection performance of weighted template plus verifier with various soft thresholds.}
\label{fig:soft-threshold}
\end{figure}

\paragraph*{Soft threshold}
We tried the verifier with different soft thresholds. Fig. \ref{fig:soft-threshold} shows the evaluation results of using different soft thresholds and its effect on the performance of the weighted template approach. As it is illustrated, the performance is improved by increasing the soft threshold until score of $0.8$, and it is decreased when we use $0.9$ as soft threshold. Thus in overall comparison results Fig. \ref{fig:overal-results}, we apply $th_{soft}=0.8$ for verifying the resulting detections which are extracted by the upper-body detector with weighted template. We get the best performance by combining the depth-based detector and the appearance based verifier.

\begin{figure*}[t]
  \centering%
  \setlength{\subfigcapskip}{0pt}
  \setlength{\subfigbottomskip}{0pt}
  
  \begin{tabularx}{\textwidth}{XXXX} 
  \hspace{2cm}jointDeep\cite{Ouyang13ICCV} & \hspace{1.5cm}Weighted Template & \hspace{0.6cm}Weighted Template + Verifier \\
  \end{tabularx}
\\[6pt]
  	\includegraphics[width=.33\textwidth]{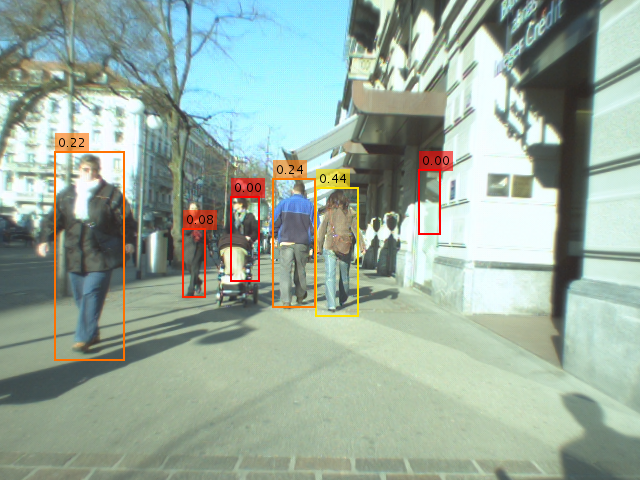}\hspace{1pt}%
  	\includegraphics[width=.33\textwidth]{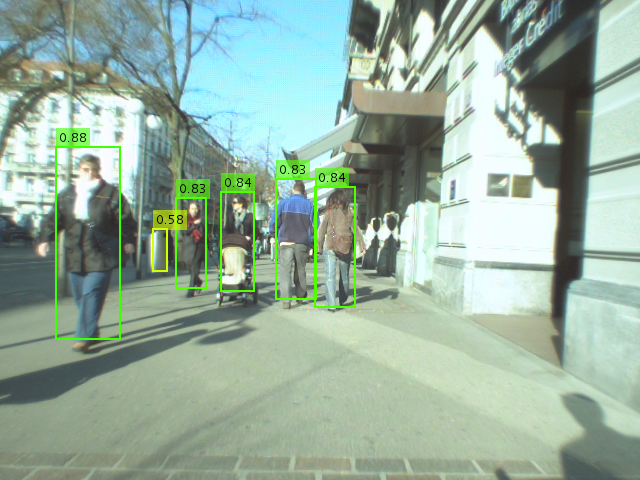}\hspace{1pt}%
  	\includegraphics[width=.33\textwidth]{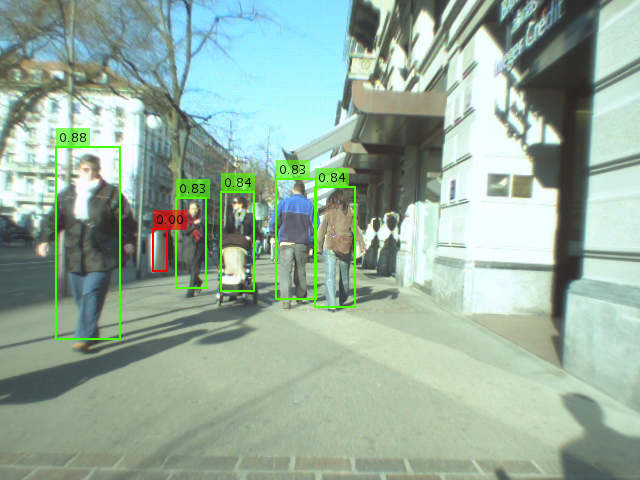}\hspace{1pt}%

  	\includegraphics[width=.33\textwidth]{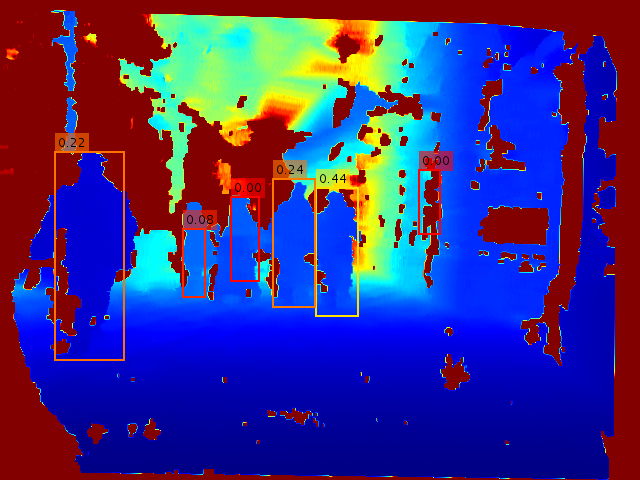}\hspace{1pt}%
  	\includegraphics[width=.33\textwidth]{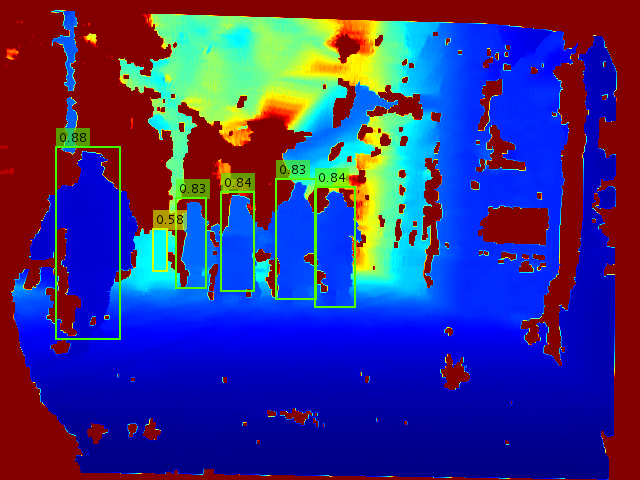}\hspace{1pt}%
  	\includegraphics[width=.33\textwidth]{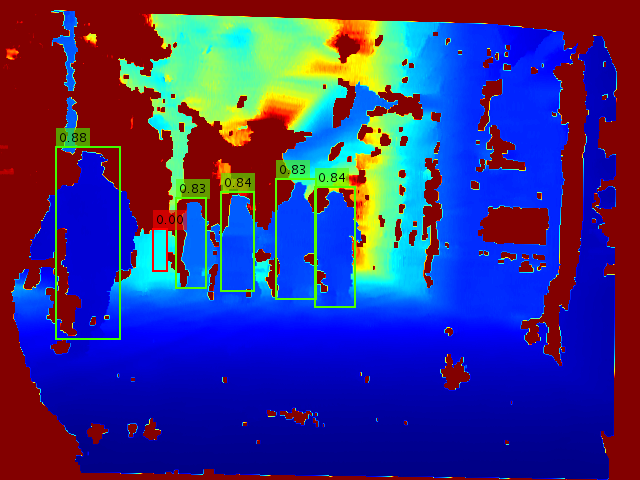}\hspace{1pt}%
\\[6pt]
	\includegraphics[width=.33\textwidth]{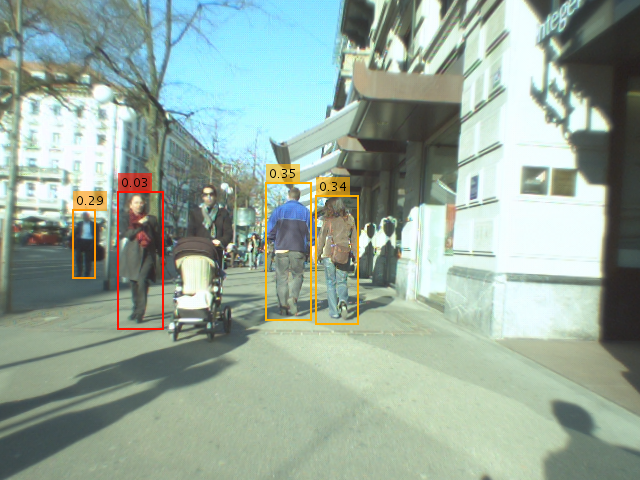}\hspace{1pt}%
	\includegraphics[width=.33\textwidth]{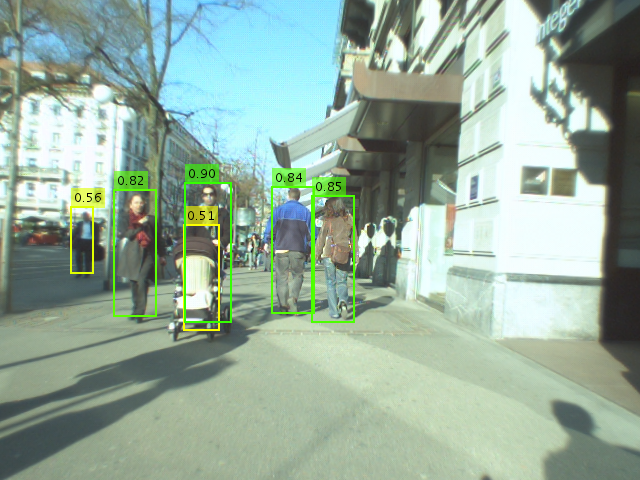}\hspace{1pt}%
	\includegraphics[width=.33\textwidth]{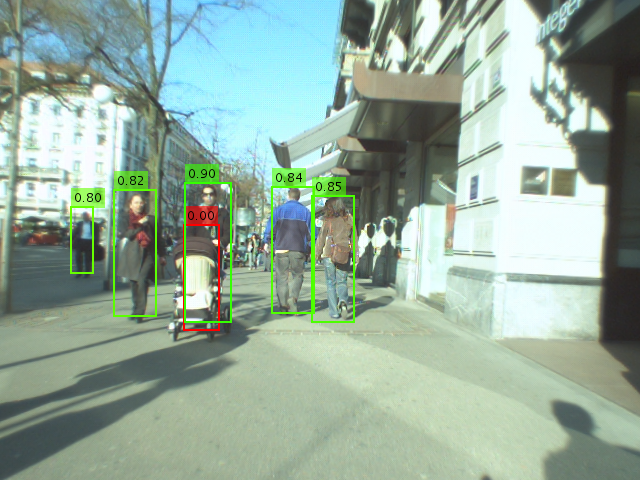}\hspace{1pt}%

	\includegraphics[width=.33\textwidth]{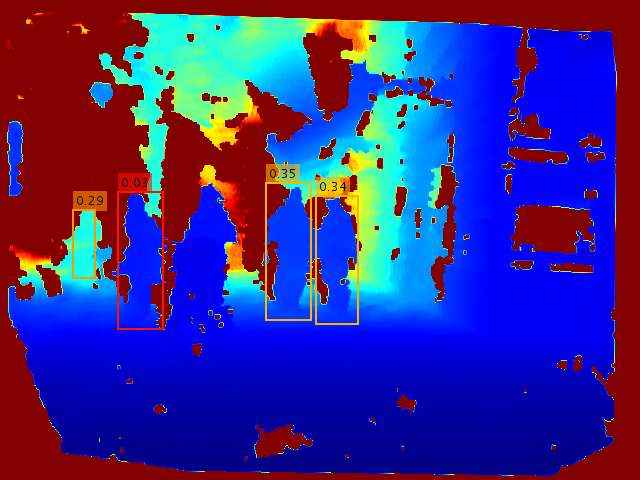}\hspace{1pt}%
	\includegraphics[width=.33\textwidth]{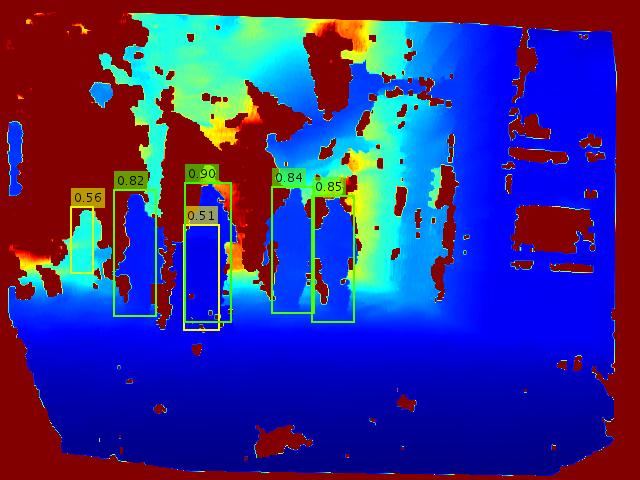}\hspace{1pt}%
	\includegraphics[width=.33\textwidth]{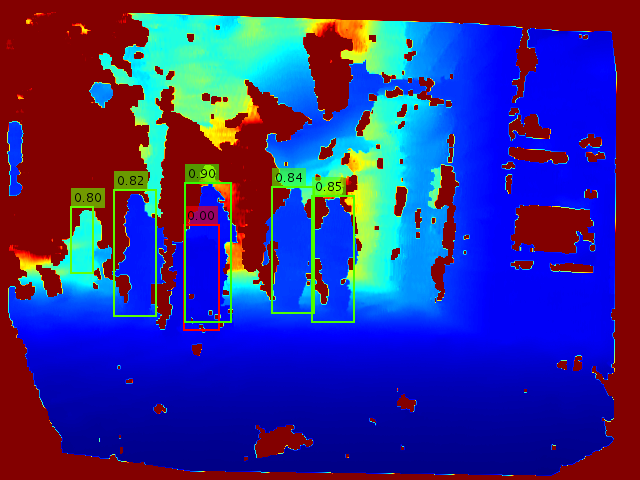}\hspace{1pt}%

    \includegraphics[width=0.7\textwidth]{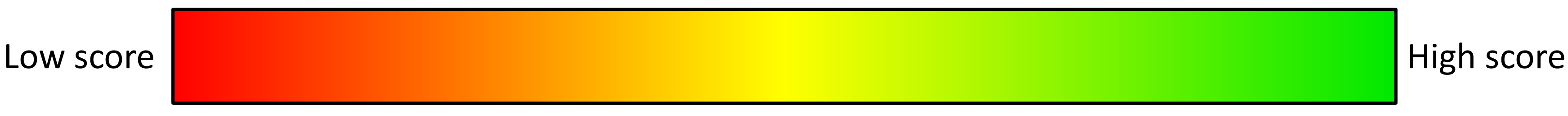}
 \caption[]{Qualitative results on the ETH Zurich Sunnyday dataset \cite{Ess09ICRA}. The resulting detections are illustrated on the RGB image and the depth-map of two challenging frames for jointDeep (left column), weighted template upper-body detector (middle column) and weighted template upper-body detector with jointDeep as verifier(right column). Our approach can resolve some challenging scenarios such as (1) objects with similar depth-map with upper-body (trash can in top frame and child stroller in bottom frame) and (2) noisy depth map of the pedestrians in far distance from the camera (the man walking in left side of bottom frame).}
  \label{fig:qual_res}
\end{figure*}

\begin{figure*}[t]
  \centering%
  \setlength{\subfigcapskip}{0pt}
  \setlength{\subfigbottomskip}{0pt}
  
  \begin{tabularx}{\textwidth}{>{\setlength\hsize{1\hsize}\centering}X>{\setlength\hsize{1\hsize}\centering}X>{\setlength\hsize{1\hsize}\centering}X}
  jointDeep\cite{Ouyang13ICCV} & Weighted Template & Weighted Template + Verifier
  \end{tabularx}
\\[6pt]
  	\includegraphics[width=.33\textwidth]{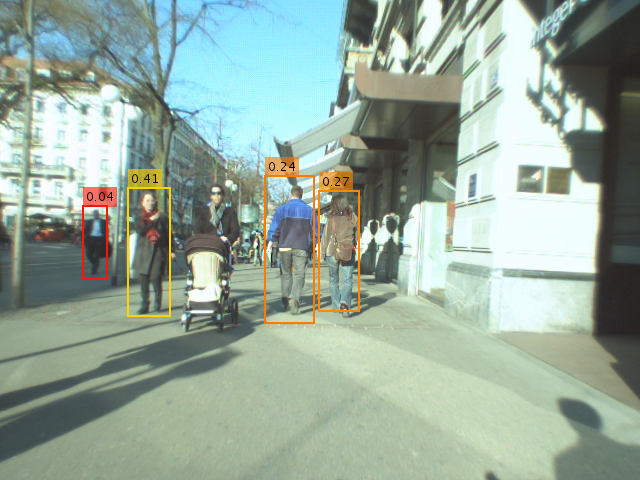}\hspace{1pt}%
  	\includegraphics[width=.33\textwidth]{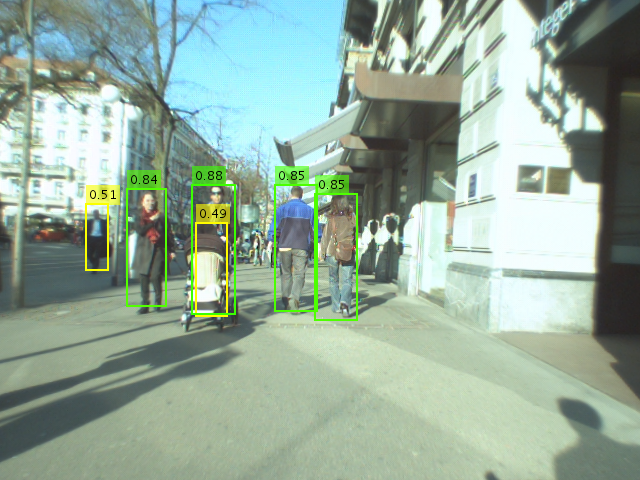}\hspace{1pt}%
  	\includegraphics[width=.33\textwidth]{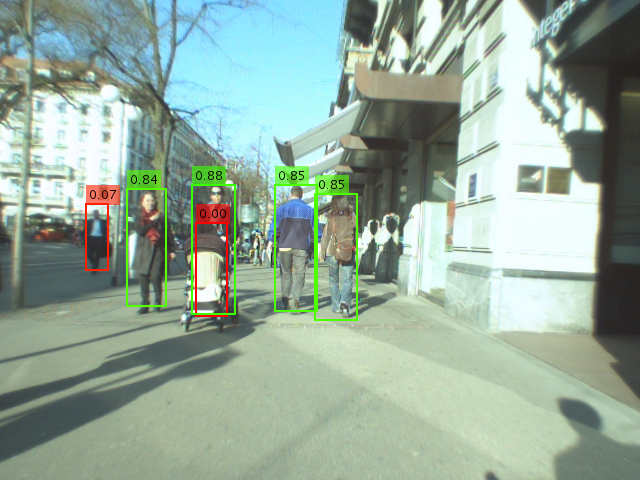}\hspace{1pt}%

  	\includegraphics[width=.33\textwidth]{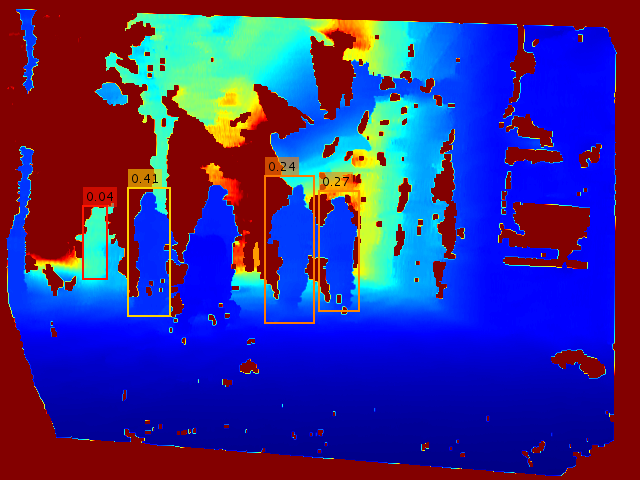}\hspace{1pt}%
  	\includegraphics[width=.33\textwidth]{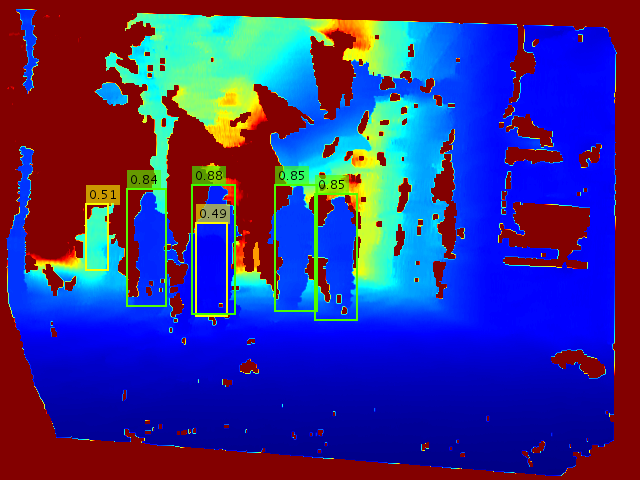}\hspace{1pt}%
  	\includegraphics[width=.33\textwidth]{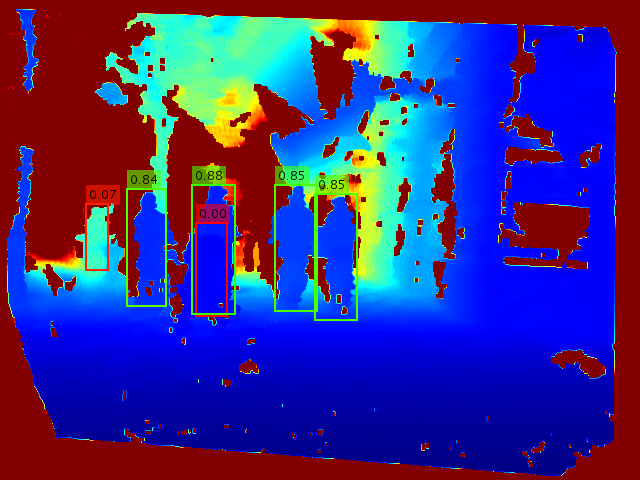}\hspace{1pt}%

 \caption[]{Failure case. In some cases, the verifier reduces the score of the detected pedestrians. As it is shown in this figure, the man walking in the left side of the scene is detected by the weighted template approach with the score of $0.51$. Although, we expect to refine this detection by using the appearance based verifier, but it fails and reduces the to $0.07$.}
  \label{fig:qual_fail}
\end{figure*}

In Fig. \ref{fig:qual_res}, we compare the resulting detections of jointDeep detector, weighted template and their combination \footnote{More qualitative results are provided in the supplementary material.}. One of the most challenging issues in the depth-based pedestrian detectors is the limitation of the depth-map when we have similar objects to the upper-body. As illustrated in Fig. \ref{fig:qual_res}, the trash can and child stroller are detected by the weighted template with scores of $0.58$ and $0.51$ accordingly. Since their scores is lower than $th_{soft}=0.80$, these detections are verified with the appearance based detector and their scores decreased to $\approx 0$. Another challenging issue is noise of the depth-map for far away pedestrians. In third and fourth rows of Fig. \ref{fig:qual_res}, we show this issue is handled for the man walking in the left side of the scene. We also show a failure case in Fig. \ref{fig:qual_fail} which may be eliminated by using a better appearance based detector as the verifier.

\subsection{Computational Performance}
Table \ref{tab:runtime} shows the run-time for each component of our pipeline. The evaluations are done on a machine with Intel i7-4790K CPU (using only single core of CPU). The verifying phase is done as a post processing step. The publicly available MATLAB code of the jointDeep detector \cite{Ouyang13ICCV} is used as the verifier. The verifier can be optimized to  reduce the run-time. The frame rate of the whole pipeline (including verifier) is 22 \textit{fps} and without verifier it is 45 \textit{fps}.

\begin{table*}[t]
\centering
 \caption{The average computational run-time which is needed for detecting pedestrians inside a one frame.}
\begin{tabular}{|l|c|c|c|c|}
  \hline
  &\textbf{Single-template\cite{HosseiniICRA14}}&\textbf{Weighted-template}&\textbf{Multiple orientation-based}&\textbf{Multiple distance-based}\\
  \hhline{|=|=|=|=|=|}
  upper-body detector (ms) & 7 & 8 & 10 & 7\\
  \hline
  Ground plane estimation (ms) & 4 & 4 & 4 & 4 \\
  \hline
  ROI processing (ms) & 10 & 10 & 10 & 10 \\
  \hline
  Verifier: jointDeep \cite{Ouyang13ICCV} (ms) & 22 & 22 & 22 & 22 \\
  \hhline{|=|=|=|=|=|}
  \textbf{Total without verifier (ms)} & 21 & 22 & 24 & 21 \\
  \hline
  \textbf{Total with verifier (ms)} & 43 & 44 & 46 & 43 \\
  \hline
\end{tabular}
 \label{tab:runtime}
\end{table*}

\section{CONCLUSION}
In this paper, we present a depth-based template matching pedestrian detection which uses an appearance based verifier for refining the detections. We proposed three methods for training the depth-based template. Weighted template is introduced for weighing the distance measurement and emphasizing the importance of the reliable regions inside the sliding window. Using the orientation-based multiple templates, we can handle different orientations of the upper-body of human during detection. By utilizing the distance-based multiple templates, we handle various levels of  detail of the depth-map for the upper-bodies corresponding to the pedestrians with different distances from the camera. Furthermore, an appearance based verifier is used to compensate the limitations of the depth-maps. Our proposed approach runs in real-time and outperforms the state-of-the-art pedestrian detectors.
\bibliographystyle{IEEEtran}
\bibliography{IEEEabrv,root}

\end{document}